\documentclass{article}

\usepackage[preprint]{neurips_2024}

\usepackage[utf8]{inputenc} 
\usepackage[T1]{fontenc}    
\usepackage{hyperref}       
\usepackage{url}            
\usepackage{booktabs}       
\usepackage{amsfonts}       
\usepackage{nicefrac}       
\usepackage{microtype}      
\usepackage{xcolor}         
\usepackage{amsmath}
\usepackage{graphicx}
\usepackage{wrapfig}
\usepackage{amssymb}
\usepackage[utf8]{inputenc}
\usepackage{booktabs}  
\newcommand{\red}[1]{\textcolor{red}{#1}}

\title{The Crystal Ball Hypothesis in diffusion models: Anticipating object positions from initial noise}

\author{%
  Yuanhao Ban\\
  UCLA\\
  \texttt{banyh2000@gmail.com} \\
  \And
  Ruochen Wang \\
  UCLA\\
  \texttt{ruocwang@g.ucla.edu} \\
  \And
  Tianyi Zhou \\
  UMD \\
  \texttt{tianyi@umd.edu}
  \And
  Boqing Gong \\
  Google \\
  \texttt{bgong@google.com} \\
  \And
  Cho-Jui Hsieh \\
  UCLA \\
  \texttt{chohsieh@cs.ucla.edu} \\
  \And
  Minhao Cheng \\
  PSU \\
  \texttt{mmc7149@psu.edu}
}

\begin{document}

\maketitle

\begin{abstract}
 
  Diffusion models have achieved remarkable success in text-to-image generation tasks; however, the role of initial noise has been rarely explored. In this study, we identify specific regions within the initial noise image, termed trigger patches, that play a key role for object generation in the resulting images. Notably, these patches are ``universal'' and can be generalized across various positions, seeds, and prompts. To be specific, extracting these patches from one noise and injecting them into another noise leads to object generation in targeted areas. We identify these patches by analyzing the dispersion of object bounding boxes across generated images, leading to the development of a posterior analysis technique.  Furthermore, we create a dataset consisting of Gaussian noises labeled with bounding boxes corresponding to the objects appearing in the generated images and train a detector that identifies these patches from the initial noise. To explain the formation of these patches, we reveal that they are outliers in Gaussian noise, and follow distinct distributions through two-sample tests. Finally, we find the misalignment between prompts and the trigger patch patterns can result in unsuccessful image generations. The study proposes a reject-sampling strategy to obtain optimal noise, aiming to improve prompt adherence and positional diversity in image generation.

\end{abstract}

\section{Introduction}
\vspace{-4pt}
  \begin{figure}[ht]
      \centering      \includegraphics[width=\textwidth]{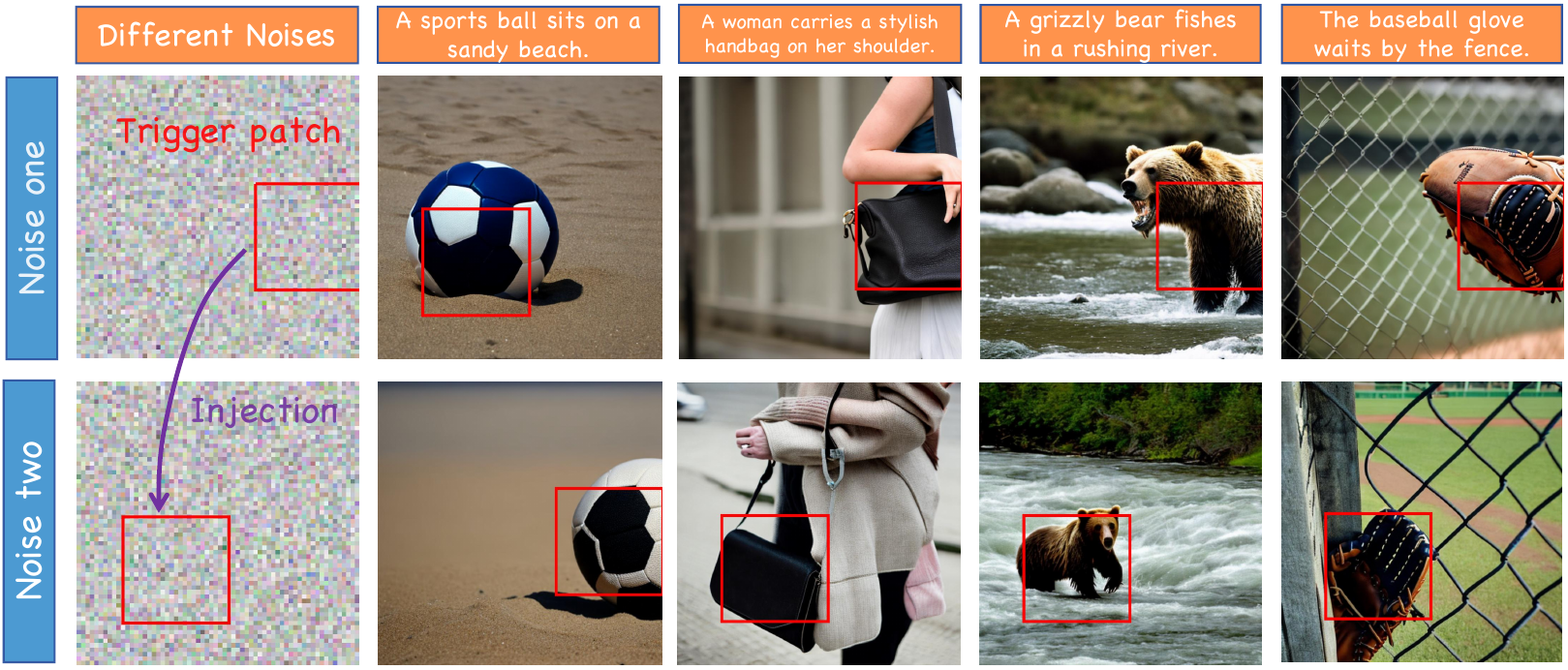}
      \vspace{-20pt}
      \caption{Illustration: The first row shows images generated from one seed. And we can identify the ``trigger patch'' located in the red box that tend to induce the generation of the object. If we inject the trigger patch into another noise, there will be objects in the position of the injection place in the images generated by the mixed noise.}
      \label{fig:Illu}
      \vspace{-14pt}
  \end{figure}

 In recent years, diffusion models have revolutionized the field of text-to-image generation~\citep{saharia2022photorealistic,rombach2022high,dhariwal2021diffusion,nichol2021glide,ho2022classifier}. However, these models often fail to accurately adhere to the prompts, frequently generating objects with specific positions or attributes regardless of the input text~\citep{chefer2023attend,hertz2022prompt,wang2022diffusiondb}. Although various methods aimed at enhancing control over the generation process have been introduced, including modifying the denoising process~\citep{balaji2022ediff}, manipulating cross-attention layers~\citep{hertz2022prompt,feng2022training}, and retraining models using layout-image pairs~\citep{zheng2023layoutdiffusion,zhang2023adding,voynov2023sketch}, an important question still left answered that why is the generation process so difficult to control?

 In this paper, we show that  Gaussian noise in the diffusion process plays a crucial role in image generation. Specifically, we discover the existence of {\bf trigger patches} -- distinct patches in the noise space that trigger the generation of objects in the diffusion model. By moving the trigger patch to a different position, the corresponding object will likely move to that location. Furthermore, this trigger effect also exists across various prompts—the same trigger patch can initiate the generation of different objects, depending on the given prompt.  These phenomena are illustrated in Figure~\ref{fig:Illu}: when replacing the target patch within \textbf{another initial noise} with the trigger patch, the injection position would generate an object. Identifying the location of trigger patches can provide insights into where objects will be generated without running the diffusion process.  Moreover, moving/removing trigger patches can achieve certain image editing effects. 

 {\it How can we find trigger patches?} To give a quantitative way to identify trigger patches, we first propose a posterior analysis approach based on the following intuition: given a specific initial noise,  if \textbf{all the objects are confined to a specific location regardless of the prompts}, there must be a trigger patch in the corresponding position. Hence, we define ``trigger entropy'' based on the variance of coordinates of detected objects in the generated images. {\it But can we detect trigger patches without actually running a diffusion process?} We answer this question affirmatively by training a ``trigger patch detector'', which functions similarly to an object detector but operates in the noise space. Our trigger patch detector achieves an mAP50 of 0.325 on the validation set\footnote{Faster R-CNN~\cite{ren2015faster} achieves a mAP$_{50}$ of 0.596 on COCO validation set~\cite{lin2014microsoft}.}, revealing that trigger patches possess distinct properties. 

 {\it So what makes trigger patches special?} We hypothesize that the trigger patches are \textbf{outliers} within the Gaussian noise. We perform a two-sample test comparing the trigger patches to randomly selected noise patches and confirm that they follow distinct distributions. In particular, more effective trigger patches show greater divergence from the standard Gaussian distribution. To further support our claim, we design some handcrafted trigger patches shifted from the original distribution and validate that they effectively trigger objects at corresponding positions generated images.

Finally, we demonstrate two applications of trigger patches. For scenarios where positional bias is desired, such as when location information is specified in the prompt, we show that finding noises with trigger patches at the target location can significantly boost the generation success rate from 57.08\% to 83.64\%. Conversely, for applications where location bias is undesirable, such as when location diversity of generated objects is preferred, we demonstrate that using our proposed detector to "purify" the noise can significantly increase generation diversity.



%

\vspace{-6pt}
\section{Existance of Trigger Patches and How to Find Them}
\vspace{-6pt}
\subsection{Prelimiaries}
\vspace{-6pt}
\paragraph{Denoising Diffusion Probabilistic Models(DDPM)} DDPM~\cite{ho2020denoising} is a kind of generative model, which achieves great performance on high-quality image synthesis. The inference process initiates with the standard Gaussian noise $x_T$ and iteratively applies the model to progressively denoise it back towards the natural image $x_0$ that subjects to the real data distribution.
\vspace{-8pt}
\paragraph{Classifer-free guidance for conditional generation}
Text-to-image diffusion models incorporate classifier-free context information into the reverse diffusion process via cross-attention layers. Specifically, at each sampling step, the denoising error is calculated by adjusting the conditional error with an unconditional error, factored by a guidance strength.

\begin{wrapfigure}{r}{0.33\textwidth}
  \centering
  \vspace{-4pt}  \includegraphics[width=0.33\textwidth]{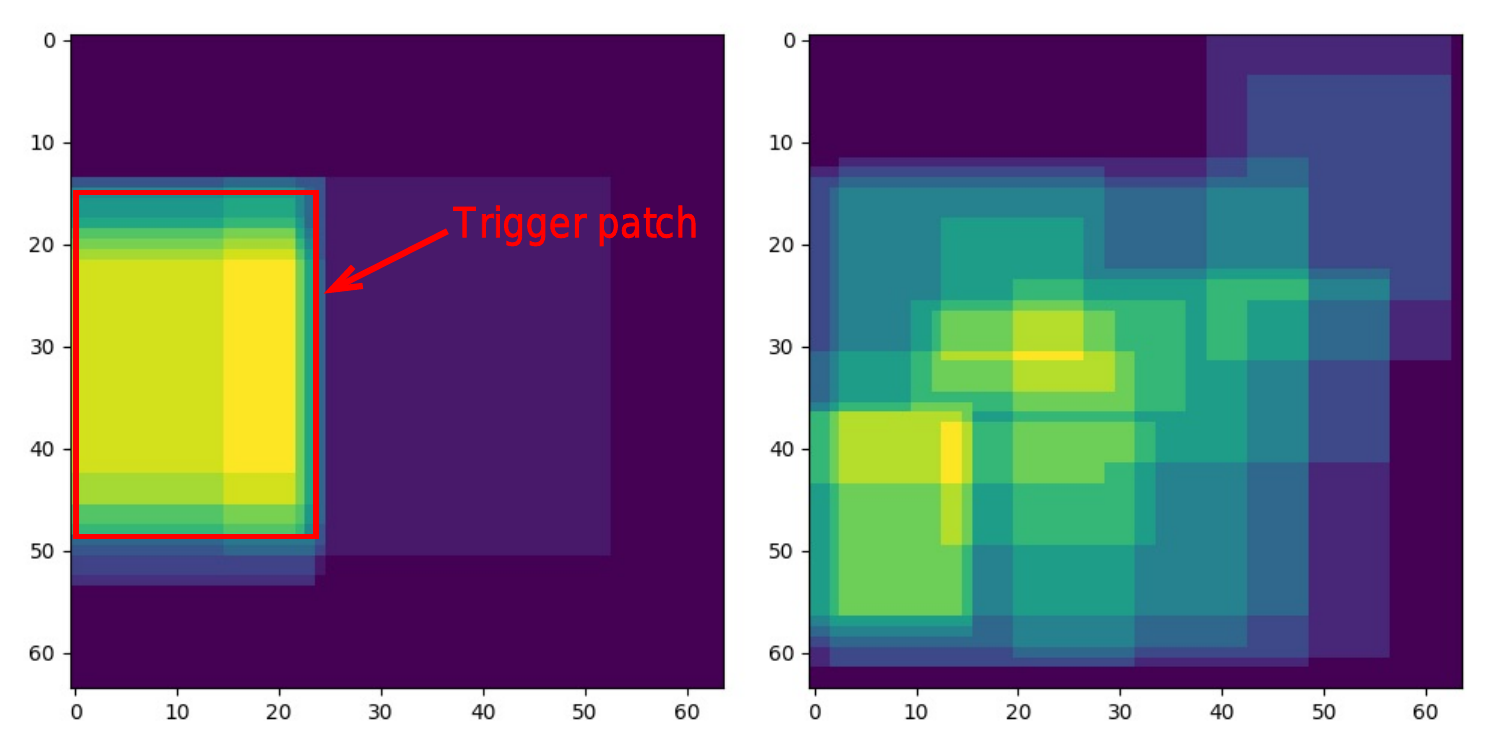}
  \vspace{-20pt}
  \caption{Heatmap of generated objects on two different noises. The left one has a trigger patch while the right one does not. }
  \label{fig:heatmap}
  \vspace{-12pt}
\end{wrapfigure}
\vspace{-4pt}
\subsection{Definition of the trigger patch} 
Where will a diffusion model position an object in the generated images? In this section, we show that {\it the initial noise takes an important role in generated objects' location}. Specifically, we identify regions within the initial noise, termed ``{\bf trigger patches},'' that largely determine the object location. Formally, a trigger patch is a patch in the noise space with the following properties: (1) Triggering Effect: When it presents in the initial noise $Z_0$, the trigger patch consistently induces object generation at its corresponding location; (2) Universality Across Prompts: The same trigger patch can trigger the generation of various objects, depending on the given prompt.

Fig~\ref{fig:heatmap}
 illustrates the existence of trigger patches and demonstrates their universality across prompts. For a fixed initial noise, we generate 25 images with various prompts, and then detect the object in each generated image. Summing the object masks and normalizing the result yields a heatmap, where each pixel represents the probability of an object appearing at that location.
 Fig~\ref{fig:heatmap} demonstrates the heatmap for two different noises. For the left one, the bounding boxes are consistently aligned on the left side, indicating the presence of a trigger patch that consistently leads to object generation at that location. Conversely, the right heatmap has no such an obvious pattern, suggesting no trigger patch in that particular noise. 

\vspace{-4pt}
\subsection{Identifying trigger patch: a posterior analysis}
\begin{figure}[t]
      \centering      \includegraphics[width=\textwidth]{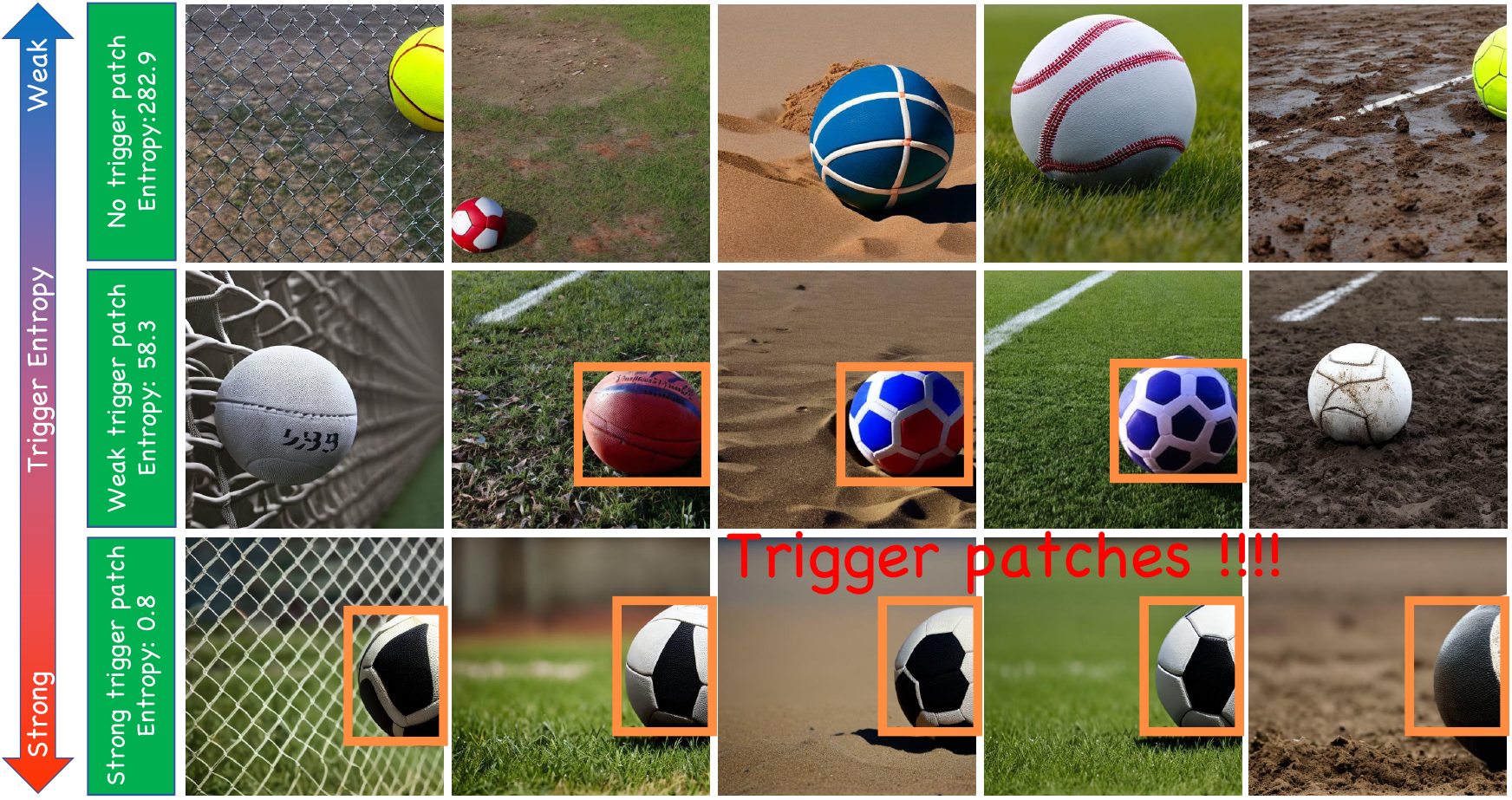}
      \vspace{-20pt}
      \caption{Illustration: Each row shows five images generated from one seed. The patch in the bottom row has the strongest effectiveness in inducing the generation of the object, while the objects in the top row are dispersed. So there must exist a strong trigger patch in the orange bounding box in the bottom row noise.}
     \label{fig:triggerentropy}
     \vspace{-12pt}
  \end{figure}

\vspace{-4pt}
In this subsection, we propose a posterior analysis to identify and define trigger patches. Intuitively, we generate a bunch of images using the same noise and calculate a posterior metric called ``trigger entropy'' to measure the degree of dispersion of the object bounding boxes in the generated images. A lower entropy indicates a higher likelihood that a patch is a trigger patch, as shown in Fig~\ref{fig:triggerentropy}. We also conduct experiments to assess the efficacy of patches at various levels of trigger entropy in generating objects across different categories and initial noises.

\begin{figure}[h]
    \centering
    \begin{minipage}{0.315\textwidth}
      \centering
    \includegraphics[width=\textwidth]{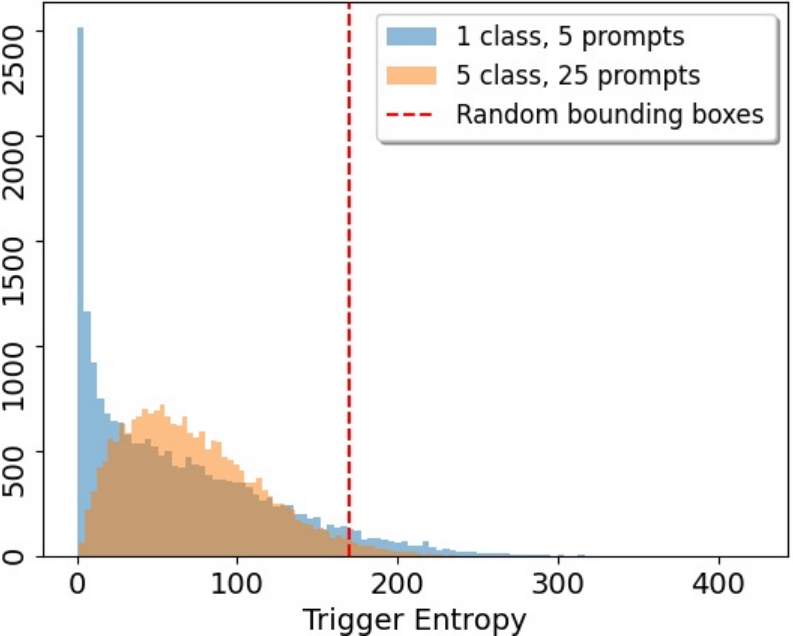}
      \vspace{-20pt}
      \caption{Trigger Entropy Distribution on the created dataset. Randomly selected bounding boxes as a baseline.}
      \label{fig:variance}
    \end{minipage}\hfill
    \begin{minipage}{0.315\textwidth}
      \centering
      \includegraphics[width=\textwidth]{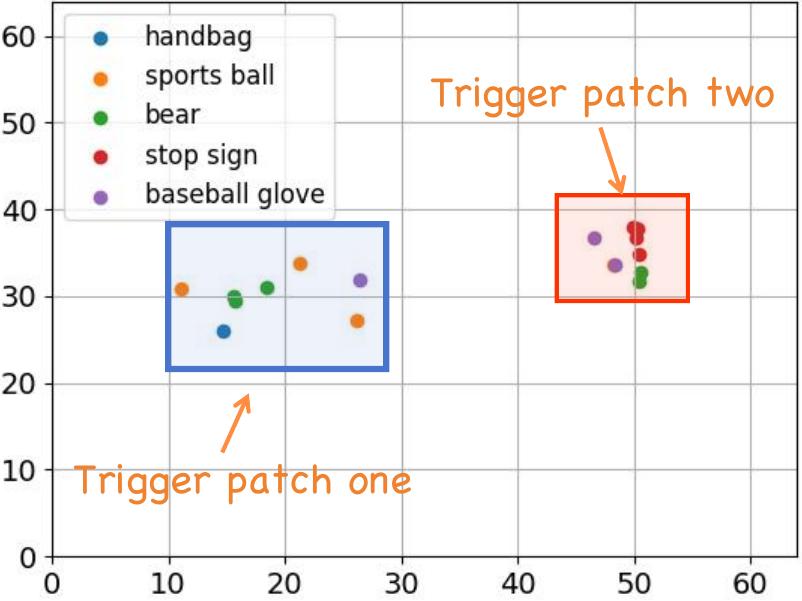}
      \caption{Object Distribution: For one noise, we plot the scatter to see where the center of the bounding boxes are dispersed.}
      \label{fig:cluster}
    \end{minipage}\hfill
        \begin{minipage}{0.345\textwidth}
      \centering
      \includegraphics[width=\textwidth]{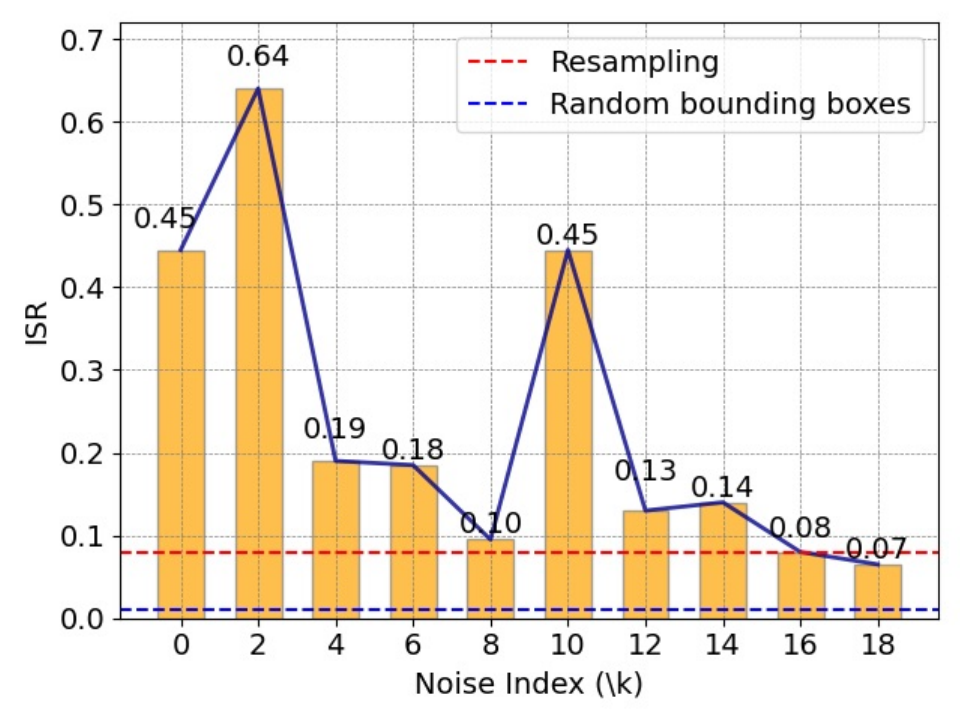}
      \vspace{-18pt}
      \caption{Trigger Injection: We sort the 20,000 trigger patches by Trigger entropy and do injection experiments on every 2000 noise.}
      \label{fig:triggerinjection}
    \end{minipage}
\end{figure}

\vspace{-8pt}
\paragraph{Dataset} We generate a large-scale dataset consisting of initial noises paired with the bounding boxes of objects in the corresponding images. Initially, we select five classes from the COCO dataset: ``stop sign, bear, sports ball, handbag, and apple.'' We then prompt ChatGPT to generate five unique sentences for each class, ensuring each sentence features only one object to prevent duplicates. We sample 20,000 Gaussian noises and use diffusion models to generate images based on these noises and prompts. A pre-trained object detector from MMDetection~\cite{chen2019mmdetection} is then applied to identify the object bounding boxes, which are resized from the image space to the latent space. Only the bounding box with the highest score and correct object label is retained for each image. Please refer to Appendix~\ref{app:crafting} for more details.
\vspace{-8pt}
\paragraph{Metric} After obtaining the coordinates of the bounding boxes $\{({x}_1,{y}_1,{x}_2,{y}_2)\}$, we can define a posterior metric about the effectiveness of the regions in inducing objects, termed ``trigger entropy''. For each initial noise, the more concentrated the boxes are, the smaller trigger entropy it has. To avoid the bias of the object size, we compute the center point of each object bounding box ${x}_c = \frac{{x}_1 + {x}_2}{2}$ and ${y}_c = \frac{{y}_1 + {y}_2}{2}$. Then the entropy of the patch can be defined based on the variance of the box centers:
\begin{align} 
\label{eq:entropy}
\mathcal{H}(X) = \frac{1}{2} \left( \frac{1}{n} \sum_{i=1}^n (x_{c_i} - \bar{x}_c)^2 + \frac{1}{n} \sum_{i=1}^n (y_{c_i} - \bar{y}_c)^2 \right)
\end{align}
where $\: \bar{x}_c = \frac{1}{n} \sum_{i=1}^n x_{c_i},\: \bar{y}_c = \frac{1}{n} \sum_{i=1}^n y_{c_i}$, and $i$ is the box index.
So, the more entropy a region has, the less effective it is in determining the objects' locations. 
\vspace{-8pt}
\paragraph{Dataset Statistics} We then compute the trigger entropy of all the 20,000 noises and plot the histograms in Fig~\ref{fig:variance}. We first study a trivial case, where we compute the trigger entropy based on the boxes of one class, ``sports ball'', and five prompts about this class. Then for a more general one, we compute the trigger entropy based on all 5 classes and 25 prompts to see if the metric can generalize well across various prompts and classes. As shown in the Figure, nearly 10\% noises in the first group have an entropy near 0. In other words, for nearly 2,000 noises, sports balls always show up in almost the same place across the five images generated from an initial noise, indicating a strong trigger patch over that place. 

\vspace{-4pt}
To further explore the characteristics of the trigger patches within an initial noise, we randomly select a noise from the dataset and plot the center points of the detected bounding boxes in Fig~\ref{fig:cluster}. We can see that it is possible that multiple trigger patches exist within a single noise sample. Consequently, the posterior metric, which is based on the variance of all bounding boxes, may fail to capture the trigger patches in such scenarios, which will be further addressed in Sec~\ref{sec:detector}. Moreover, the trigger patch does not depend on the specific prompt --- both trigger patches in the figure can generate baseball gloves (purple points) and bears (green points). These experiments confirm that our trigger patches are \textbf{universal}.
\vspace{-8pt}
\paragraph{Trigger Injection} To verify these patches with low trigger entropy are indeed trigger patches, we design an experiment to test if these patches can still take effect when injected into another noise. In particular, we replace an arbitrary patch in a noise map with the trigger patch and then use the blended noise for generation. A successful injection is characterized by the detected bounding box occupying more than 75 percent of the trigger patch region. Then we define the injection success rate (ISR) as the ratio of successful injection cases to the total number of cases. We sort the patches in our dataset by trigger entropy and select one every two thousand for this test. We do the injection experiment with 200 new random noises. There are two baselines as follows. 1) Resampling: Resample Gaussian noise for the target region maintaining the same mean and variance. 2) Random: Select a random patch within a source noise, which might overlap with the trigger patch in the source noise. Figure~\ref{fig:triggerentropy} illustrates that the Injection Success Rates (ISRs) of most trigger patches exceed those of the Resampling (0.08) and Random (0.01) baselines, confirming the efficacy of the trigger patches identified by our posterior entropy. Additionally, a lower trigger entropy correlates with a higher ISR, suggesting that the trigger patches with low entropy are particularly effective in determining object locations.

\vspace{-8pt}
\paragraph{Trigger-Prompt Interaction} In the previous paragraph, we examine the object position information contained in trigger patches. The analysis thus far raises a subsequent question: What happens when the prompts also contain positional information, and how do the prompt and noise interact with each other when they have aligned and contradicted object positions? To explore this, we design experiments to investigate \textbf{the interplay between the trigger patches and the prompts}. In particular, we design prompts with the format of ``a {coco class name} on the right side''. Then we select three noises: 1) Aligned: with a strong trigger patch on the right, 2) Contradicted: with a strong trigger patch on the left, and 3) Dispersed: with trigger patches dispersed throughout the image. After viewing the generated images, we divided them into four groups: 1) Aligned: The position of the object is aligned with the prompt guidance (On the right side). 2) Contradicted: The position of the object contradicts with the prompt guidance (On the left side). 3) Duplicated: the generated image has two objects on both sides. 4) Hard to judge (Occupying the entire picture, failed generation, or in the middle of the image). 
\begin{table}[ht]
\centering
\caption{Diversity Results}
\label{tab:interaction}
\vspace{-5pt}
\begin{tabular}{l|c|c|c|c}
\hline\hline
Noise &  Aligned (\%) & Contradicted (\%) & Duplicated (\%) & Hard to judge (\%)\\
\hline
Aligned & 63.5  & 6.3 & 6.3 & 23.9 \\
\hline
Contradicted & 35.0 & 32.5 & 8.8 & 23.7 \\
\hline
Dispersed & 25.0 & 10.0 & 11.3 & 53.7 \\
\hline
\end{tabular}
\end{table}

The main findings are as follows. Table~\ref{tab:interaction} demonstrates that the trigger patches significantly influence the final positions of objects. When the prompt and trigger patches are contradictory, 32.5\% of objects adhere to the trigger patches, while only 35.0\% follow the prompt's guidance, highlighting the prevalence of failed generations due to \textbf{conflicts between the prompt and initial noise about where to position objects.} Notably, when the prompt and trigger patches are aligned, 63.5\% of objects are accurately positioned. These results indicate the potential to manipulate trigger patches for controlled image generation, as demonstrated in Section~\ref{sec:promptfollowing}. 

\vspace{-4pt}
\subsection{Detecting trigger patches directly from noise}\label{sec:detector}
\vspace{-4pt}
The above posterior method requires generating images to identify trigger patches. Can we detect the trigger patches without running the diffusion process? To answer this question, we try to train a trigger patch detector, which functions similarly to an object detector but operates in the noise space. 

  \begin{figure}[ht]
  \centering  
  \includegraphics[width=0.85\textwidth]{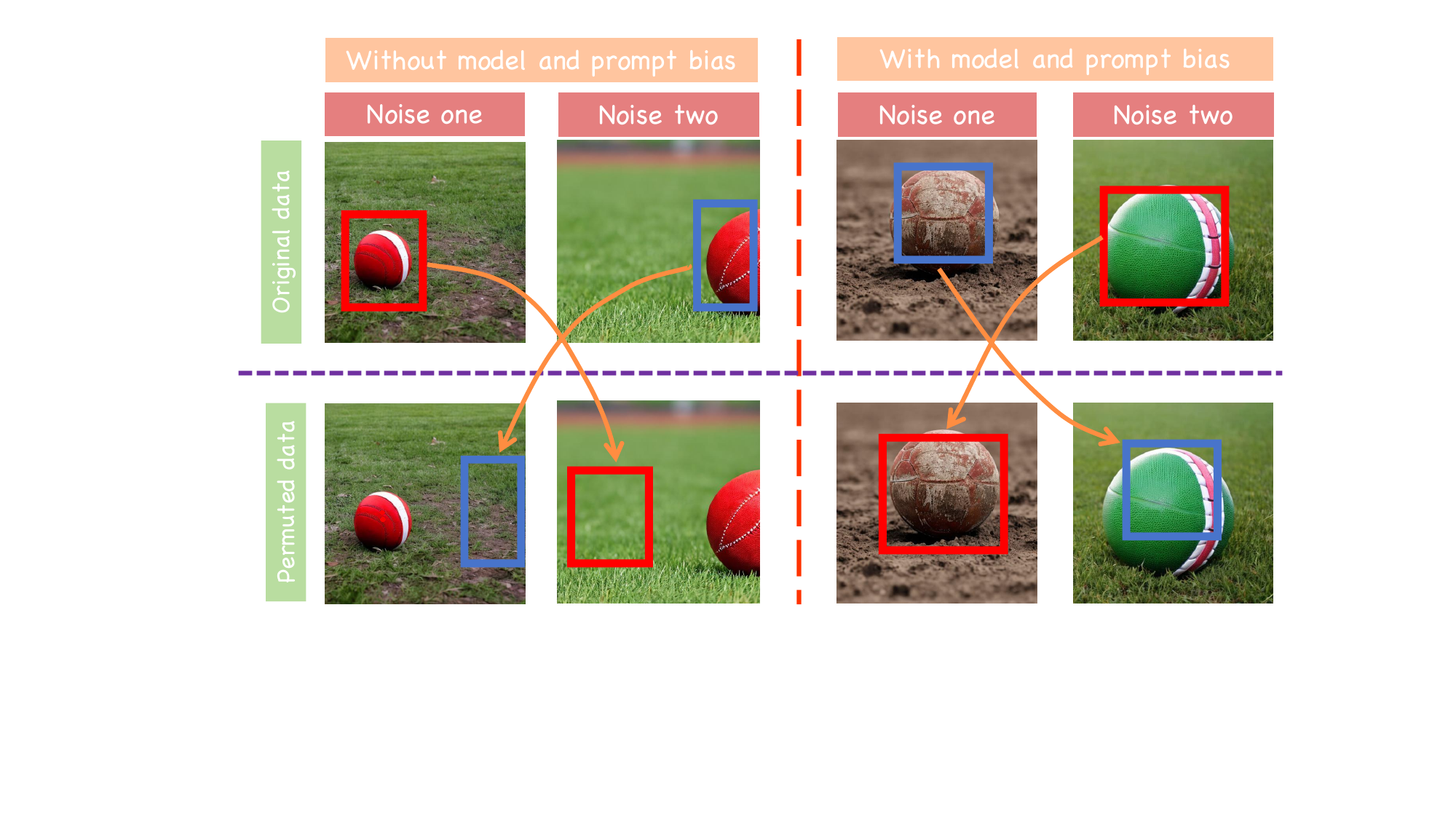}
  \vspace{-5pt}
  \caption{Permuted Data: We permuted the annotations associated with the noise samples. Initially, in the top row, Noise One was labeled with Annotation One, and Noise Two with Annotation Two. After permutation, shown in the bottom row, Noise One is now labeled with Annotation Two, and Noise Two with Annotation One. Ideally, in scenarios with a strong bias, permuting the annotations should have minimal or no impact on the final results. This is because, in both permuted and unpermuted datasets, the bounding boxes remain centrally located within the image.}
  \vspace{-8pt}
  \label{fig:permuted}
  \end{figure}
  \vspace{-8pt}
\noindent

\paragraph{Dataset} We divide 20,000 random noises into a training set (17,500), a validation set (1,000), and a test set (1,500). Subsequently, we modify the trigger patch annotations (obtained by using the COCO object detector~\cite{li2020generalized} in the image space) to create four distinct datasets, the details of which are presented in Table~\ref{tab:dataset}. The Augmented dataset uses bounding boxes of all the 25 prompts, while the Restricted only uses those of the class ``sports ball''. In the Class-Specific dataset, the model is required to output the class labels of the trigger patch. Additionally, we craft a dataset with permuted annotations to serve as a baseline to mitigate the effects of model and prompt biases. In particular, model positional bias, often caused by an unbalanced training dataset with all the objects appearing in one place, can lead to generated objects frequently appearing in specific image areas, typically the center. As Fig~\ref{fig:permuted} shows, if model bias is predominant, such that all bounding boxes are centralized, the training outcomes on the Permuted dataset should align with those from the Restricted dataset.
\begin{table}[ht]
\centering
\caption{Dataset Description}
\label{tab:dataset}
\vspace{-7pt}
\begin{tabular}{l|c|c|c|c}
\hline\hline
Dataset & Classes & Annotations per noise & Output class labels & Permuted annotations \\
\hline
Restricted & Sports ball & 5 = 1 $\times$ 5 & $\times$ & $\times$ \\ \hline
Augmented & All & 25 = 5 $\times$ 5 & $\times$ & $\times$ \\ \hline
Class-Specific & All & 25 = 5 $\times$ 5 & $\checkmark$ & $\times$ \\ \hline
Permuted & Sports ball & 5 = 1 $\times$ 5 & $\times$ & $\checkmark$ \\ 
\hline
\end{tabular}
\end{table}
\vspace{-16pt}


\noindent
\paragraph{Results} We utilize the MMDetection repository~\cite{chen2019mmdetection} to train our trigger patch detector. Please refer to Appendix~\ref{app:training} for more details. The results can be seen in Table~\ref{tab:detector}. Our detector on Restricted achieves the mAP$_{50}$ of 0.325, surpassing the Permuted baseline by 0.124. Such evidence eliminates the influence of the model bias and prompt bias, verifying that the model has learned to locate the trigger patches from initial noises. Notably, the results on the Augmented dataset show that the trigger patches can generalize across different text prompts and provide a foundation for the applications of our detector in various scenarios. Meanwhile, the detector on dataset Class-Specific shows degenerated performance, indicating that the trigger patches are ignorant of the classes. In other words, these patches can only determine where objects are generated, not what to generate. This confirms our finding in Fig~\ref{fig:cluster} that the same patch can trigger different objects, depending on what is given in the prompt. 
\begin{table}[ht]
\centering
\vspace{-4pt}
\caption{Main results: Training a detector to anticipate object positions from initial noise.}
\label{tab:detector}
\vspace{-7pt}
\begin{tabular}{c|c|c|c|c}
\hline\hline
Dataset & Restricted & Augmented & Class-Specific & Randomized \\
\hline
mAP$_{50}$ & 0.325 & 0.333 & 0.091 & 0.201 \\
\hline
\end{tabular}
\end{table}
\vspace{-12pt}
\section{What contributes to the formation of trigger patches}
\vspace{-4pt}
In this part, we propose a potential explanation for the existence of trigger patches. While previous work~\cite{guo2024initno,sun2024spatial,zheng2023layoutdiffusion} mainly focuses on the cross-attention maps and analyzes the interactions between specific noise, prompts, and models, we hypothesize that \textbf{these trigger patches are typical outliers in the sampled noise}, and have minimal correlation with prompts. Furthermore, we propose a method to synthesize trigger patches with great success. 

\vspace{-4pt}
\subsection{Two-sample test} 
\vspace{-4pt}
\begin{wrapfigure}{r}{0.35\textwidth}
  \centering
  \vspace{-12pt}  \includegraphics[width=0.35\textwidth]{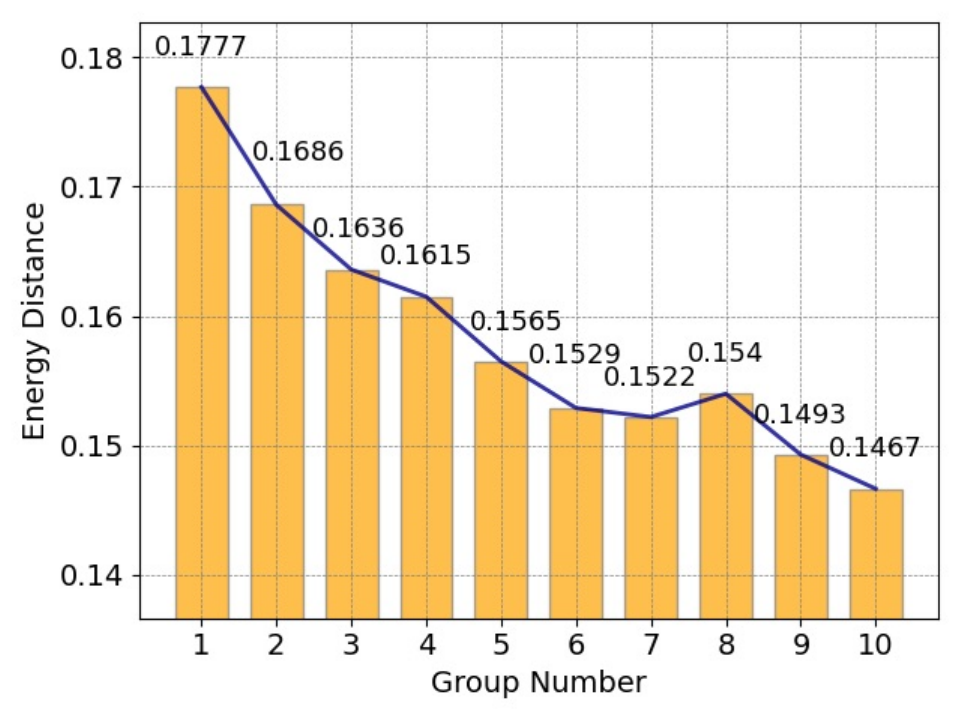}
  \vspace{-25pt}
  \caption{Energy Distance}
  \label{fig:distance}
  \vspace{-12pt}
\end{wrapfigure}
We perform an energy-based two-sample test, which measures the distance between the distributions of two samples by computing the sum of pairwise Euclidean distances among all sample points~\cite{szekely2013energy}. Meanwhile, it provides a $p$-value, which statistically quantifies the extent to which the distributions differ from each other. A low $p$-value (typically less than 0.05) suggests that the two sets of samples are not subjected to the same distribution. To test our hypothesis that the trigger patches are outliers in the initial noises, we divide 20,000 noises into ten groups according to the order of Trigger Entropy. Each group has 2,000 noises, and the first group has the lowest Trigger Entropy. Then we refer to the annotations, compute the center point of each bounding box, and extract the trigger patches with a fixed size 24$\times$24 around the center. Following these steps, we obtain a list of trigger patches for each group. We then create ten negative groups by randomly selecting a patch of 24$\times$24 in the same noise. Additionally, we craft a control group by randomly selecting patches of the same size. The results show that all the $p$-values between trigger patch groups and negative prompts are 0.0, while the $p$-value between the negative prompts and the control group is 0.938. Hence, the trigger patches and those random patches in the negative groups are from different distributions. Furthermore, we examine the relationship between the energy distance of trigger patches from negative patches and the trigger entropy, as illustrated in Figure~\ref{fig:distance}. This figure reveals a clear negative correlation: the greater the deviation of a patch from the original distribution, the more likely it is to be a trigger patch characterized by low trigger entropy. This observation again supports our hypothesis that these trigger patches are outliers within the noise.
\vspace{-6pt}
\subsection{Hand-crafted trigger patches}
\vspace{-4pt}
If the trigger patches are indeed outliers, it may be feasible to construct certain signals that deviate from a normal Gaussian distribution to serve as artificially created trigger patches. To this end, we conduct experiments to test artificial trigger patches with various non-Gaussian signals.
\begin{enumerate}
  \item Natural trigger patches: We extract the trigger patches with the lowest trigger entropy from the 20,000 noise samples.
  \item Resampling baseline: We resample Gaussian noise maintaining the same mean and variance to establish a baseline.
  \item Random baseline: We randomly select 25 sets, each consisting of four numbers, from a range of 512. These numbers are interpreted as the coordinates for the final detected bounding boxes, upon which we compute entropy. This baseline is designed to simulate scenarios where each patch has an equal probability of generating an object.  
  \item Shifted Gaussian: We sample Gaussian noises with altered standard deviation (std) so the resultant noise follows a different Gaussian from the diffusion models' default noises.
  \item Sine Function: We create sinusoidal noise patches by firstly using a function that applies a sine transformation to each coordinate axis $(x, y, z)$ and then summing up the values. We then add them to an initial noise via interpolation as follows,
\begin{align} 
\label{eq:designed}
\overline{P_{(x,y,z)}} = \sin(\theta)\cdot\left[\sin(\frac{2\pi x}{l_x}) + \sin(\frac{2\pi y}{l_y}) + \sin(\frac{2\pi z}{l_z})\right] + \cos(\theta)\cdot P_{(x,y,z)},
\end{align}
where $P_{(x,y,z)}$ is the pixel value of the original patch at position $(x,y,z)$, $l_x$, $l_y$ and $l_z$ are the widths of the trigger patch, and $\theta$ is the interpolation parameter. In our case, $l_x$ $l_y$ and $l_z$ are 24, 24 and 4, respectively.
\end{enumerate}

The results are displayed in Table~\ref{tab:handcrafted}. Natural trigger patches achieved a success rate of 44.5\%, which exceeds the rates of randomly selected bounding boxes and the resampling of the region by 42.5\% and 36.0\%, respectively. 
Notably, Sin Function patches yield comparable results when moderate interpolation parameters are set, achieving an ISR of 49\%. When the interpolation parameter $\theta$ is set at 0.15$\cdot\frac{\pi}{2}$, it achieves a significantly high ISR of 81.0\%. However, this setting also causes image distortion. Interestingly, sampling Gaussian noise with a larger standard deviation appears to be more effective in inducing trigger patches than using a smaller one, verifying that these trigger patches might be considered outliers. 

\begin{table}[htbp]
\centering
\vspace{-4pt}
\caption{Main results on the injection experiments. The higher the ISR, the better. The left major column shows the ISRs for the baseline methods: Random, Resampling, and Natural. The middle major column shows the results for the shifted Gaussian method with different standard deviations(STD). The right major column shows the results for the Sine Function with different interpolation weights.}
\label{tab:handcrafted}
\vspace{-7pt}
\begin{tabular}{@{}lc|cc|cc@{}} 
\toprule
 \multicolumn{2}{c|}{Baselines} & \multicolumn{2}{c|}{Shifted Gaussian} & \multicolumn{2}{c}{Sine Function} \\
\cmidrule(r){1-2} \cmidrule(r){3-4} \cmidrule(r){5-6}
     & ISR(\%) & STD          & ISR(\%) &      $\theta$($\cdot\frac{\pi}{2}$)    & ISR(\%) \\ 
\midrule
Random            & 1.0         & 0.8         &  8.5           & 0.08    & 33.5   \\
Resampling            &  8.5        & 1.2       & 29.0                & 0.10     & 49.0     \\
Natural            & 44.5            & 1.5      & 90.0           & 0.15     & 81.0      \\
\bottomrule
\end{tabular}
\end{table}
\vspace{-14pt}

\section{Applications}
\vspace{-4pt}

The ability to detect trigger patches enhances control over object locations in generated images, opening the door to numerous applications. Here, we demonstrate two such applications: 1) In scenarios where the prompt does not have positional information, we aim to increase the positional diversity of the generated images. 2) Conversely, when the prompt includes explicit positional guidance, our goal is to ensure that the generated images follow the provided directions. It is important to note that manipulating trigger patches alone may not deliver state-of-the-art results. Our main objective is to demonstrate that even such a straightforward approach can achieve reasonable performance, highlighting the significant role that trigger patches play in the generation process.

\vspace{-4pt}
\subsection{Enhanced location diversity by removing trigger patches}
\vspace{-4pt}
\paragraph{Background} 
Researchers have observed various biases in diffusion models, including gender bias~\cite{luccioni2024stable} and color bias~\cite{orgad2023editing}. However, position bias, which is the tendency for objects to consistently appear in the same locations across different generations, has received minimal attention. This type of bias has significant implications, particularly in synthetic data generation using diffusion models. For instance, when generating an automobile dataset, a model exhibiting position bias might consistently place roadblocks in the left-bottom corner of the image, leading to skewed data. Moreover, such a bias can affect the diversity of generated images in applications like ChatGPT, where similar images are produced for varied prompts due to the presence of trigger patches with low trigger entropy in the initial noise. 
\vspace{-8pt}
\paragraph{Method}To mitigate this, we have developed a method involving the use of the detector for reject sampling. This process begins with the detection of trigger patches in the initial noise. If the confidence scores of the bounding boxes exceed a predefined threshold, the region within the box is flagged for regeneration. Our approach is designed to ensure that the noise used for generation is ``pure'' and has no strong trigger patches that could confine object placement.
\vspace{-8pt}
\paragraph{Baselines} Although researchers have studied the positional bias in GAN-based generative models~\cite{choi2021toward}, no counterpart works have been found on the diffusion models. Focusing on the role of initial noise, \textbf{Initno}~\cite{guo2024initno} may be the most related one, though it needs the specific prompt for each generation process to take effect. Furthermore, we establish another \textbf{Control} group by generating images using standard methods without reject-sampling. Additionally, we select \textbf{Random} bounding boxes of size 24$\times$24 from the initial noise. This approach simulates a scenario where all pixels have an equal probability of generating an object, which is our target.
\vspace{-8pt}
\paragraph{Metrics and Coniguration} We calculate the average entropy for the bounding boxes in these groups to assess the impact of our method on diversity and positional bias. The entropy is calculated as outlined in Equation \eqref{eq:entropy}, with higher values indicating greater diversity. Our experimental setup includes generating images from 10 random prompts described by ChatGPT, followed by image generation using diffusion models. Please refer to the Appendix~\ref{app:diversity} for more details. The objects' bounding boxes are identified using a COCO detector, and entropy calculations are conducted on the bounding boxes with confidence scores over 0.75. We set the rejection threshold for the confidence score of the detector as 0.6 and conducted our experiments using 1000 different seeds. 
\begin{table}[ht]
\centering
\vspace{-4pt}
\caption{Diversity Results}
\vspace{-5pt}
\label{tab:diversity}
\begin{tabular}{c|c|c|c|c}
\hline\hline
Methods &  Control & Initno & Ours & Random \\
\hline
Entropy  &  135.97 & 139.89 & 171.84 & 170.64 \\
\hline
\end{tabular}
\end{table}

As we can see from the table, images generated by our methods show great diversity in position, with an entropy of 171.84, surpassing the control group by 31.95, very close to the pure random outputs, which is 170.64. Meanwhile, Initno shows bad performance with an entropy of 139.89, which is behind ours by a large margin.

\vspace{-6pt}
\subsection{Better prompt following by injecting trigger patches} \label{sec:promptfollowing}
\vspace{-4pt}
\paragraph{Background and Methods} Research indicates that diffusion models often struggle to adhere to the positional information specified in prompts. This issue is typically due to the appearance of trigger patches in unintended locations. For instance, a prominent trigger patch on the left side of an image can conflict with a prompt like ``a dog on the left", resulting in generation failures. To address this, we propose a reject-sampling technique to obtain an optimal initial noise where trigger patches align with the prompt requirements even before generation. Specifically, we continue to resample noises until the center point of the bounding box, which scores highest according to our detector, is positioned within the area targeted by the prompt.
\vspace{-8pt}
\paragraph{Baselines and Settings} Our methodology is compared against Initno~\cite{guo2024initno} and a baseline model that does not utilize reject-sampling. To assess our method's effectiveness, we design 10 prompts incorporating the words \textbf{right} or \textbf{left} and generate 500 images for each prompt. We employ a pre-trained COCO detector to verify whether the objects are correctly positioned as indicated by the prompt and calculate the guidance success rate (GSR) based on the proportion of correctly positioned objects to the total number of cases. Please refer to Appendix~\ref{app:injecting} for more details.
\vspace{-8pt}
\paragraph{Results} Table ~\ref{tab:guidance} illustrates that our method, relying solely on the initial noise, achieves an impressive guidance success rate (GSR) of 83.64\%.
\begin{table}[ht]
\centering
\caption{Diversity Results}
\label{tab:guidance}
\vspace{-7pt}
\begin{tabular}{c|c|c|c}
\hline\hline
Methods &  Control & Initno & Ours \\
\hline
GSR(\%)  &  57.08 & 61.08 & 83.64 \\
\hline
\end{tabular}
\vspace{-12pt}
\end{table}

\section{Related Work}
\vspace{-8pt}
\paragraph{Initial noise} In DDIM~\cite{song2020denoising}, the final image is determined by the model parameters, the text prompt, and the initial noise. While the model parameters and text prompts have been extensively researched, the initial noise remains relatively under-explored. \cite{mao2023guided} first highlights that initial noises exhibit distinct preferences for certain layouts and analyze this phenomenon through the lens of cross-attention. \cite{sun2024spatial} utilized an inverted reference image with finite inversion steps to incorporate valuable spatial awareness, leading to a new framework for controllable image generation. \cite{guo2024initno} attributed failed generation cases to 'bad initial noise' and proposed optimizing the initial noise via cross-attention to achieve better results.

Building on these studies, our research focuses on the role of initial noise. Unlike previous works that qualitatively suggest the presence of positional information in initial noise, we provide evidence for the existence of trigger patches that tend to generate specific objects by defining a metric to quantitatively assess the effectiveness of these patches. Additionally, we propose that these trigger patches are universal and capable of generalizing across different positions, prompts, and background noises. While prior research primarily examines noise from a cross-attention perspective, we discover that the intriguing properties might stem from the statistical characteristics of these patches, which are outliers within the initial noise. Finally, we identify significant applications in enhancing positional diversity. While previous studies mainly addressed biases related to gender, color, and texture, our findings reveal that noise with strong trigger patches can introduce positional bias.
\vspace{-8pt}
\paragraph{Types of objection detection} Object detection is a fundamental computer vision technique that identifies and locates objects in digital images and videos. Various detection networks have been developed for different application scenarios, such as static images~\cite{ren2015faster}, video frames~\cite{wojke2017simple}, depth images~\cite{reading2021categorical}, point clouds~\cite{qi2017pointnet}, sequential data~\cite{lea2017temporal}, and multi-channel images~\cite{chen2016deep}.
However, we propose a novel and counter-intuitive approach: using pure Gaussian noise as the input, the model is trained to learn from these noises and extract patches with intriguing properties. To the best of our knowledge, this is the first study to propose such an approach.

\vspace{-8pt}
\section{Limitations}
\vspace{-8pt}
We have not thoroughly explored scenarios where one noise sample contains multiple trigger patches and have limited our dataset to five classes and 25 prompts due to computational constraints. A more diverse, large-scale dataset is planned for future research.
\vspace{-8pt}
\section{Conclusion} 
\vspace{-8pt}
In this study, we have uncovered specific trigger patches within the initial noise that are likely to induce object generation. We subsequently developed a detector capable of identifying these trigger patches from the initial noise prior to the generation process. We characterized these trigger patches as outliers within Gaussian noise and conducted experiments to validate this classification. Finally, we show two potential applications for our detector, both of which have demonstrated excellent performance. We leave more applications for future work.

\bibliographystyle{plainnat}
\bibliography{main.bib}

\begin{thebibliography}{31}
\providecommand{\natexlab}[1]{#1}
\providecommand{\url}[1]{\texttt{#1}}
\expandafter\ifx\csname urlstyle\endcsname\relax
  \providecommand{\doi}[1]{doi: #1}\else
  \providecommand{\doi}{doi: \begingroup \urlstyle{rm}\Url}\fi

\bibitem[Balaji et~al.(2022)Balaji, Nah, Huang, Vahdat, Song, Zhang, Kreis, Aittala, Aila, Laine, et~al.]{balaji2022ediff}
Yogesh Balaji, Seungjun Nah, Xun Huang, Arash Vahdat, Jiaming Song, Qinsheng Zhang, Karsten Kreis, Miika Aittala, Timo Aila, Samuli Laine, et~al.
\newblock ediff-i: Text-to-image diffusion models with an ensemble of expert denoisers.
\newblock \emph{arXiv preprint arXiv:2211.01324}, 2022.

\bibitem[Chefer et~al.(2023)Chefer, Alaluf, Vinker, Wolf, and Cohen-Or]{chefer2023attend}
Hila Chefer, Yuval Alaluf, Yael Vinker, Lior Wolf, and Daniel Cohen-Or.
\newblock Attend-and-excite: Attention-based semantic guidance for text-to-image diffusion models.
\newblock \emph{ACM Transactions on Graphics (TOG)}, 42\penalty0 (4):\penalty0 1--10, 2023.

\bibitem[Chen et~al.(2019)Chen, Wang, Pang, Cao, Xiong, Li, Sun, Feng, Liu, Xu, et~al.]{chen2019mmdetection}
Kai Chen, Jiaqi Wang, Jiangmiao Pang, Yuhang Cao, Yu~Xiong, Xiaoxiao Li, Shuyang Sun, Wansen Feng, Ziwei Liu, Jiarui Xu, et~al.
\newblock Mmdetection: Open mmlab detection toolbox and benchmark.
\newblock \emph{arXiv preprint arXiv:1906.07155}, 2019.

\bibitem[Chen et~al.(2016)Chen, Jiang, Li, Jia, and Ghamisi]{chen2016deep}
Yushi Chen, Hanlu Jiang, Chunyang Li, Xiuping Jia, and Pedram Ghamisi.
\newblock Deep feature extraction and classification of hyperspectral images based on convolutional neural networks.
\newblock \emph{IEEE transactions on geoscience and remote sensing}, 54\penalty0 (10):\penalty0 6232--6251, 2016.

\bibitem[Choi et~al.(2021)Choi, Lee, Jeong, and Yoon]{choi2021toward}
Jooyoung Choi, Jungbeom Lee, Yonghyun Jeong, and Sungroh Yoon.
\newblock Toward spatially unbiased generative models.
\newblock \emph{arXiv preprint arXiv:2108.01285}, 2021.

\bibitem[Dhariwal and Nichol(2021)]{dhariwal2021diffusion}
Prafulla Dhariwal and Alexander Nichol.
\newblock Diffusion models beat gans on image synthesis.
\newblock \emph{Advances in neural information processing systems}, 34:\penalty0 8780--8794, 2021.

\bibitem[Feng et~al.(2022)Feng, He, Fu, Jampani, Akula, Narayana, Basu, Wang, and Wang]{feng2022training}
Weixi Feng, Xuehai He, Tsu-Jui Fu, Varun Jampani, Arjun Akula, Pradyumna Narayana, Sugato Basu, Xin~Eric Wang, and William~Yang Wang.
\newblock Training-free structured diffusion guidance for compositional text-to-image synthesis.
\newblock \emph{arXiv preprint arXiv:2212.05032}, 2022.

\bibitem[Guo et~al.(2024)Guo, Liu, Cui, Li, Yang, and Huang]{guo2024initno}
Xiefan Guo, Jinlin Liu, Miaomiao Cui, Jiankai Li, Hongyu Yang, and Di~Huang.
\newblock Initno: Boosting text-to-image diffusion models via initial noise optimization.
\newblock \emph{arXiv preprint arXiv:2404.04650}, 2024.

\bibitem[Hertz et~al.(2022)Hertz, Mokady, Tenenbaum, Aberman, Pritch, and Cohen-Or]{hertz2022prompt}
Amir Hertz, Ron Mokady, Jay Tenenbaum, Kfir Aberman, Yael Pritch, and Daniel Cohen-Or.
\newblock Prompt-to-prompt image editing with cross attention control.
\newblock \emph{arXiv preprint arXiv:2208.01626}, 2022.

\bibitem[Ho and Salimans(2022)]{ho2022classifier}
Jonathan Ho and Tim Salimans.
\newblock Classifier-free diffusion guidance.
\newblock \emph{arXiv preprint arXiv:2207.12598}, 2022.

\bibitem[Ho et~al.(2020)Ho, Jain, and Abbeel]{ho2020denoising}
Jonathan Ho, Ajay Jain, and Pieter Abbeel.
\newblock Denoising diffusion probabilistic models.
\newblock \emph{Advances in neural information processing systems}, 33:\penalty0 6840--6851, 2020.

\bibitem[Lea et~al.(2017)Lea, Flynn, Vidal, Reiter, and Hager]{lea2017temporal}
Colin Lea, Michael~D Flynn, Rene Vidal, Austin Reiter, and Gregory~D Hager.
\newblock Temporal convolutional networks for action segmentation and detection.
\newblock In \emph{proceedings of the IEEE Conference on Computer Vision and Pattern Recognition}, pages 156--165, 2017.

\bibitem[Li et~al.(2020)Li, Wang, Wu, Chen, Hu, Li, Tang, and Yang]{li2020generalized}
Xiang Li, Wenhai Wang, Lijun Wu, Shuo Chen, Xiaolin Hu, Jun Li, Jinhui Tang, and Jian Yang.
\newblock Generalized focal loss: Learning qualified and distributed bounding boxes for dense object detection.
\newblock \emph{Advances in Neural Information Processing Systems}, 33:\penalty0 21002--21012, 2020.

\bibitem[Lin et~al.(2014)Lin, Maire, Belongie, Hays, Perona, Ramanan, Doll{\'a}r, and Zitnick]{lin2014microsoft}
Tsung-Yi Lin, Michael Maire, Serge Belongie, James Hays, Pietro Perona, Deva Ramanan, Piotr Doll{\'a}r, and C~Lawrence Zitnick.
\newblock Microsoft coco: Common objects in context.
\newblock In \emph{Computer Vision--ECCV 2014: 13th European Conference, Zurich, Switzerland, September 6-12, 2014, Proceedings, Part V 13}, pages 740--755. Springer, 2014.

\bibitem[Luccioni et~al.(2024)Luccioni, Akiki, Mitchell, and Jernite]{luccioni2024stable}
Sasha Luccioni, Christopher Akiki, Margaret Mitchell, and Yacine Jernite.
\newblock Stable bias: Evaluating societal representations in diffusion models.
\newblock \emph{Advances in Neural Information Processing Systems}, 36, 2024.

\bibitem[Mao et~al.(2023)Mao, Wang, and Aizawa]{mao2023guided}
Jiafeng Mao, Xueting Wang, and Kiyoharu Aizawa.
\newblock Guided image synthesis via initial image editing in diffusion model.
\newblock In \emph{Proceedings of the 31st ACM International Conference on Multimedia}, pages 5321--5329, 2023.

\bibitem[Nichol et~al.(2021)Nichol, Dhariwal, Ramesh, Shyam, Mishkin, McGrew, Sutskever, and Chen]{nichol2021glide}
Alex Nichol, Prafulla Dhariwal, Aditya Ramesh, Pranav Shyam, Pamela Mishkin, Bob McGrew, Ilya Sutskever, and Mark Chen.
\newblock Glide: Towards photorealistic image generation and editing with text-guided diffusion models.
\newblock \emph{arXiv preprint arXiv:2112.10741}, 2021.

\bibitem[Orgad et~al.(2023)Orgad, Kawar, and Belinkov]{orgad2023editing}
Hadas Orgad, Bahjat Kawar, and Yonatan Belinkov.
\newblock Editing implicit assumptions in text-to-image diffusion models.
\newblock In \emph{Proceedings of the IEEE/CVF International Conference on Computer Vision}, pages 7053--7061, 2023.

\bibitem[Qi et~al.(2017)Qi, Su, Mo, and Guibas]{qi2017pointnet}
Charles~R Qi, Hao Su, Kaichun Mo, and Leonidas~J Guibas.
\newblock Pointnet: Deep learning on point sets for 3d classification and segmentation.
\newblock In \emph{Proceedings of the IEEE conference on computer vision and pattern recognition}, pages 652--660, 2017.

\bibitem[Reading et~al.(2021)Reading, Harakeh, Chae, and Waslander]{reading2021categorical}
Cody Reading, Ali Harakeh, Julia Chae, and Steven~L Waslander.
\newblock Categorical depth distribution network for monocular 3d object detection.
\newblock In \emph{Proceedings of the IEEE/CVF Conference on Computer Vision and Pattern Recognition}, pages 8555--8564, 2021.

\bibitem[Ren et~al.(2015)Ren, He, Girshick, and Sun]{ren2015faster}
Shaoqing Ren, Kaiming He, Ross Girshick, and Jian Sun.
\newblock Faster r-cnn: Towards real-time object detection with region proposal networks.
\newblock \emph{Advances in neural information processing systems}, 28, 2015.

\bibitem[Rombach et~al.(2022)Rombach, Blattmann, Lorenz, Esser, and Ommer]{rombach2022high}
Robin Rombach, Andreas Blattmann, Dominik Lorenz, Patrick Esser, and Bj{\"o}rn Ommer.
\newblock High-resolution image synthesis with latent diffusion models.
\newblock In \emph{Proceedings of the IEEE/CVF conference on computer vision and pattern recognition}, pages 10684--10695, 2022.

\bibitem[Saharia et~al.(2022)Saharia, Chan, Saxena, Li, Whang, Denton, Ghasemipour, Gontijo~Lopes, Karagol~Ayan, Salimans, et~al.]{saharia2022photorealistic}
Chitwan Saharia, William Chan, Saurabh Saxena, Lala Li, Jay Whang, Emily~L Denton, Kamyar Ghasemipour, Raphael Gontijo~Lopes, Burcu Karagol~Ayan, Tim Salimans, et~al.
\newblock Photorealistic text-to-image diffusion models with deep language understanding.
\newblock \emph{Advances in Neural Information Processing Systems}, 35:\penalty0 36479--36494, 2022.

\bibitem[Song et~al.(2020)Song, Meng, and Ermon]{song2020denoising}
Jiaming Song, Chenlin Meng, and Stefano Ermon.
\newblock Denoising diffusion implicit models.
\newblock \emph{arXiv preprint arXiv:2010.02502}, 2020.

\bibitem[Sun et~al.(2024)Sun, Li, Lin, and Zhang]{sun2024spatial}
Wenqiang Sun, Teng Li, Zehong Lin, and Jun Zhang.
\newblock Spatial-aware latent initialization for controllable image generation.
\newblock \emph{arXiv preprint arXiv:2401.16157}, 2024.

\bibitem[Sz{\'e}kely and Rizzo(2013)]{szekely2013energy}
G{\'a}bor~J Sz{\'e}kely and Maria~L Rizzo.
\newblock Energy statistics: A class of statistics based on distances.
\newblock \emph{Journal of statistical planning and inference}, 143\penalty0 (8):\penalty0 1249--1272, 2013.

\bibitem[Voynov et~al.(2023)Voynov, Aberman, and Cohen-Or]{voynov2023sketch}
Andrey Voynov, Kfir Aberman, and Daniel Cohen-Or.
\newblock Sketch-guided text-to-image diffusion models.
\newblock In \emph{ACM SIGGRAPH 2023 Conference Proceedings}, pages 1--11, 2023.

\bibitem[Wang et~al.(2022)Wang, Montoya, Munechika, Yang, Hoover, and Chau]{wang2022diffusiondb}
Zijie~J Wang, Evan Montoya, David Munechika, Haoyang Yang, Benjamin Hoover, and Duen~Horng Chau.
\newblock Diffusiondb: A large-scale prompt gallery dataset for text-to-image generative models.
\newblock \emph{arXiv preprint arXiv:2210.14896}, 2022.

\bibitem[Wojke et~al.(2017)Wojke, Bewley, and Paulus]{wojke2017simple}
Nicolai Wojke, Alex Bewley, and Dietrich Paulus.
\newblock Simple online and realtime tracking with a deep association metric.
\newblock In \emph{2017 IEEE international conference on image processing (ICIP)}, pages 3645--3649. IEEE, 2017.

\bibitem[Zhang et~al.(2023)Zhang, Rao, and Agrawala]{zhang2023adding}
Lvmin Zhang, Anyi Rao, and Maneesh Agrawala.
\newblock Adding conditional control to text-to-image diffusion models.
\newblock In \emph{Proceedings of the IEEE/CVF International Conference on Computer Vision}, pages 3836--3847, 2023.

\bibitem[Zheng et~al.(2023)Zheng, Zhou, Li, Qi, Shan, and Li]{zheng2023layoutdiffusion}
Guangcong Zheng, Xianpan Zhou, Xuewei Li, Zhongang Qi, Ying Shan, and Xi~Li.
\newblock Layoutdiffusion: Controllable diffusion model for layout-to-image generation.
\newblock In \emph{Proceedings of the IEEE/CVF Conference on Computer Vision and Pattern Recognition}, pages 22490--22499, 2023.

\end{thebibliography}

\newpage
\section*{NeurIPS Paper Checklist}

\begin{enumerate}

\item {\bf Claims}
    \item[] Question: Do the main claims made in the abstract and introduction accurately reflect the paper's contributions and scope?
    \item[] Answer: \answerYes{} 
    \item[] Justification: We promise.
    \item[] Guidelines:
    \begin{itemize}
        \item The answer NA means that the abstract and introduction do not include the claims made in the paper.
        \item The abstract and/or introduction should clearly state the claims made, including the contributions made in the paper and important assumptions and limitations. A No or NA answer to this question will not be perceived well by the reviewers. 
        \item The claims made should match theoretical and experimental results, and reflect how much the results can be expected to generalize to other settings. 
        \item It is fine to include aspirational goals as motivation as long as it is clear that these goals are not attained by the paper. 
    \end{itemize}

\item {\bf Limitations}
    \item[] Question: Does the paper discuss the limitations of the work performed by the authors?
    \item[] Answer: \answerYes{} 
    \item[] Justification: Please refer to the Sec\red{6}.
    \item[] Guidelines:
    \begin{itemize}
        \item The answer NA means that the paper has no limitation while the answer No means that the paper has limitations, but those are not discussed in the paper. 
        \item The authors are encouraged to create a separate "Limitations" section in their paper.
        \item The paper should point out any strong assumptions and how robust the results are to violations of these assumptions (e.g., independence assumptions, noiseless settings, model well-specification, asymptotic approximations only holding locally). The authors should reflect on how these assumptions might be violated in practice and what the implications would be.
        \item The authors should reflect on the scope of the claims made, e.g., if the approach was only tested on a few datasets or with a few runs. In general, empirical results often depend on implicit assumptions, which should be articulated.
        \item The authors should reflect on the factors that influence the performance of the approach. For example, a facial recognition algorithm may perform poorly when image resolution is low or images are taken in low lighting. Or a speech-to-text system might not be used reliably to provide closed captions for online lectures because it fails to handle technical jargon.
        \item The authors should discuss the computational efficiency of the proposed algorithms and how they scale with dataset size.
        \item If applicable, the authors should discuss possible limitations of their approach to address problems of privacy and fairness.
        \item While the authors might fear that complete honesty about limitations might be used by reviewers as grounds for rejection, a worse outcome might be that reviewers discover limitations that aren't acknowledged in the paper. The authors should use their best judgment and recognize that individual actions in favor of transparency play an important role in developing norms that preserve the integrity of the community. Reviewers will be specifically instructed to not penalize honesty concerning limitations.
    \end{itemize}

\item {\bf Theory Assumptions and Proofs}
    \item[] Question: For each theoretical result, does the paper provide the full set of assumptions and a complete (and correct) proof?
    \item[] Answer: \answerNA{} 
    \item[] Justification: No theories are proposed.
    \item[] Guidelines:
    \begin{itemize}
        \item The answer NA means that the paper does not include theoretical results. 
        \item All the theorems, formulas, and proofs in the paper should be numbered and cross-referenced.
        \item All assumptions should be clearly stated or referenced in the statement of any theorems.
        \item The proofs can either appear in the main paper or the supplemental material, but if they appear in the supplemental material, the authors are encouraged to provide a short proof sketch to provide intuition. 
        \item Inversely, any informal proof provided in the core of the paper should be complemented by formal proofs provided in appendix or supplemental material.
        \item Theorems and Lemmas that the proof relies upon should be properly referenced. 
    \end{itemize}

    \item {\bf Experimental Result Reproducibility}
    \item[] Question: Does the paper fully disclose all the information needed to reproduce the main experimental results of the paper to the extent that it affects the main claims and/or conclusions of the paper (regardless of whether the code and data are provided or not)?
    \item[] Answer: \answerYes{} 
    \item[] Justification: .
    \item[] Guidelines:
    \begin{itemize}
        \item The answer NA means that the paper does not include experiments.
        \item If the paper includes experiments, a No answer to this question will not be perceived well by the reviewers: Making the paper reproducible is important, regardless of whether the code and data are provided or not.
        \item If the contribution is a dataset and/or model, the authors should describe the steps taken to make their results reproducible or verifiable. 
        \item Depending on the contribution, reproducibility can be accomplished in various ways. For example, if the contribution is a novel architecture, describing the architecture fully might suffice, or if the contribution is a specific model and empirical evaluation, it may be necessary to either make it possible for others to replicate the model with the same dataset, or provide access to the model. In general. releasing code and data is often one good way to accomplish this, but reproducibility can also be provided via detailed instructions for how to replicate the results, access to a hosted model (e.g., in the case of a large language model), releasing of a model checkpoint, or other means that are appropriate to the research performed.
        \item While NeurIPS does not require releasing code, the conference does require all submissions to provide some reasonable avenue for reproducibility, which may depend on the nature of the contribution. For example
        \begin{enumerate}
            \item If the contribution is primarily a new algorithm, the paper should make it clear how to reproduce that algorithm.
            \item If the contribution is primarily a new model architecture, the paper should describe the architecture clearly and fully.
            \item If the contribution is a new model (e.g., a large language model), then there should either be a way to access this model for reproducing the results or a way to reproduce the model (e.g., with an open-source dataset or instructions for how to construct the dataset).
            \item We recognize that reproducibility may be tricky in some cases, in which case authors are welcome to describe the particular way they provide for reproducibility. In the case of closed-source models, it may be that access to the model is limited in some way (e.g., to registered users), but it should be possible for other researchers to have some path to reproducing or verifying the results.
        \end{enumerate}
    \end{itemize}

\item {\bf Open access to data and code}
    \item[] Question: Does the paper provide open access to the data and code, with sufficient instructions to faithfully reproduce the main experimental results, as described in supplemental material?
    \item[] Answer: \answerYes{} 
    \item[] Justification: We will release the code, model and dataset once accepted.
    \item[] Guidelines:
    \begin{itemize}
        \item The answer NA means that paper does not include experiments requiring code.
        \item Please see the NeurIPS code and data submission guidelines (\url{https://nips.cc/public/guides/CodeSubmissionPolicy}) for more details.
        \item While we encourage the release of code and data, we understand that this might not be possible, so “No” is an acceptable answer. Papers cannot be rejected simply for not including code, unless this is central to the contribution (e.g., for a new open-source benchmark).
        \item The instructions should contain the exact command and environment needed to run to reproduce the results. See the NeurIPS code and data submission guidelines (\url{https://nips.cc/public/guides/CodeSubmissionPolicy}) for more details.
        \item The authors should provide instructions on data access and preparation, including how to access the raw data, preprocessed data, intermediate data, and generated data, etc.
        \item The authors should provide scripts to reproduce all experimental results for the new proposed method and baselines. If only a subset of experiments are reproducible, they should state which ones are omitted from the script and why.
        \item At submission time, to preserve anonymity, the authors should release anonymized versions (if applicable).
        \item Providing as much information as possible in supplemental material (appended to the paper) is recommended, but including URLs to data and code is permitted.
    \end{itemize}

\item {\bf Experimental Setting/Details}
    \item[] Question: Does the paper specify all the training and test details (e.g., data splits, hyperparameters, how they were chosen, type of optimizer, etc.) necessary to understand the results?
    \item[] Answer: \answerYes{} 
    \item[] Justification: Please refer to the Appendix.
    \item[] Guidelines:
    \begin{itemize}
        \item The answer NA means that the paper does not include experiments.
        \item The experimental setting should be presented in the core of the paper to a level of detail that is necessary to appreciate the results and make sense of them.
        \item The full details can be provided either with the code, in appendix, or as supplemental material.
    \end{itemize}

\item {\bf Experiment Statistical Significance}
    \item[] Question: Does the paper report error bars suitably and correctly defined or other appropriate information about the statistical significance of the experiments?
    \item[] Answer: \answerNo{} 
    \item[] Justification: But we conducted experiments using various seeds to ensure the robustness and reproducibility of our results. 
    \item[] Guidelines:
    \begin{itemize}
        \item The answer NA means that the paper does not include experiments.
        \item The authors should answer "Yes" if the results are accompanied by error bars, confidence intervals, or statistical significance tests, at least for the experiments that support the main claims of the paper.
        \item The factors of variability that the error bars are capturing should be clearly stated (for example, train/test split, initialization, random drawing of some parameter, or overall run with given experimental conditions).
        \item The method for calculating the error bars should be explained (closed form formula, call to a library function, bootstrap, etc.)
        \item The assumptions made should be given (e.g., Normally distributed errors).
        \item It should be clear whether the error bar is the standard deviation or the standard error of the mean.
        \item It is OK to report 1-sigma error bars, but one should state it. The authors should preferably report a 2-sigma error bar than state that they have a 96\% CI, if the hypothesis of Normality of errors is not verified.
        \item For asymmetric distributions, the authors should be careful not to show in tables or figures symmetric error bars that would yield results that are out of range (e.g. negative error rates).
        \item If error bars are reported in tables or plots, The authors should explain in the text how they were calculated and reference the corresponding figures or tables in the text.
    \end{itemize}

\item {\bf Experiments Compute Resources}
    \item[] Question: For each experiment, does the paper provide sufficient information on the computer resources (type of compute workers, memory, time of execution) needed to reproduce the experiments?
    \item[] Answer: \answerYes{} 
    \item[] Justification: \justificationTODO{}
    \item[] Guidelines: Please refer to the Appendix.
    \begin{itemize}
        \item The answer NA means that the paper does not include experiments.
        \item The paper should indicate the type of compute workers CPU or GPU, internal cluster, or cloud provider, including relevant memory and storage.
        \item The paper should provide the amount of compute required for each of the individual experimental runs as well as estimate the total compute. 
        \item The paper should disclose whether the full research project required more compute than the experiments reported in the paper (e.g., preliminary or failed experiments that didn't make it into the paper). 
    \end{itemize}
    
\item {\bf Code Of Ethics}
    \item[] Question: Does the research conducted in the paper conform, in every respect, with the NeurIPS Code of Ethics \url{https://neurips.cc/public/EthicsGuidelines}?
    \item[] Answer: \answerYes{} 
    \item[] Justification: The research presented in our paper adheres strictly to the NeurIPS Code of Ethics. 
    \item[] Guidelines:
    \begin{itemize}
        \item The answer NA means that the authors have not reviewed the NeurIPS Code of Ethics.
        \item If the authors answer No, they should explain the special circumstances that require a deviation from the Code of Ethics.
        \item The authors should make sure to preserve anonymity (e.g., if there is a special consideration due to laws or regulations in their jurisdiction).
    \end{itemize}

\item {\bf Broader Impacts}
    \item[] Question: Does the paper discuss both potential positive societal impacts and negative societal impacts of the work performed?
    \item[] Answer: \answerNA{} 
    \item[] Justification: No societal impact.
    \item[] Guidelines:
    \begin{itemize}
        \item The answer NA means that there is no societal impact of the work performed.
        \item If the authors answer NA or No, they should explain why their work has no societal impact or why the paper does not address societal impact.
        \item Examples of negative societal impacts include potential malicious or unintended uses (e.g., disinformation, generating fake profiles, surveillance), fairness considerations (e.g., deployment of technologies that could make decisions that unfairly impact specific groups), privacy considerations, and security considerations.
        \item The conference expects that many papers will be foundational research and not tied to particular applications, let alone deployments. However, if there is a direct path to any negative applications, the authors should point it out. For example, it is legitimate to point out that an improvement in the quality of generative models could be used to generate deepfakes for disinformation. On the other hand, it is not needed to point out that a generic algorithm for optimizing neural networks could enable people to train models that generate Deepfakes faster.
        \item The authors should consider possible harms that could arise when the technology is being used as intended and functioning correctly, harms that could arise when the technology is being used as intended but gives incorrect results, and harms following from (intentional or unintentional) misuse of the technology.
        \item If there are negative societal impacts, the authors could also discuss possible mitigation strategies (e.g., gated release of models, providing defenses in addition to attacks, mechanisms for monitoring misuse, mechanisms to monitor how a system learns from feedback over time, improving the efficiency and accessibility of ML).
    \end{itemize}
    
\item {\bf Safeguards}
    \item[] Question: Does the paper describe safeguards that have been put in place for responsible release of data or models that have a high risk for misuse (e.g., pretrained language models, image generators, or scraped datasets)?
    \item[] Answer: \answerNA{} 
    \item[] Justification: No such risks.
    \item[] Guidelines:
    \begin{itemize}
        \item The answer NA means that the paper poses no such risks.
        \item Released models that have a high risk for misuse or dual-use should be released with necessary safeguards to allow for controlled use of the model, for example by requiring that users adhere to usage guidelines or restrictions to access the model or implementing safety filters. 
        \item Datasets that have been scraped from the Internet could pose safety risks. The authors should describe how they avoided releasing unsafe images.
        \item We recognize that providing effective safeguards is challenging, and many papers do not require this, but we encourage authors to take this into account and make a best faith effort.
    \end{itemize}

\item {\bf Licenses for existing assets}
    \item[] Question: Are the creators or original owners of assets (e.g., code, data, models), used in the paper, properly credited and are the license and terms of use explicitly mentioned and properly respected?
    \item[] Answer: \answerYes{} 
    \item[] Justification: We have cited the model, code, and data in our study.
    \item[] Guidelines:
    \begin{itemize}
        \item The answer NA means that the paper does not use existing assets.
        \item The authors should cite the original paper that produced the code package or dataset.
        \item The authors should state which version of the asset is used and, if possible, include a URL.
        \item The name of the license (e.g., CC-BY 4.0) should be included for each asset.
        \item For scraped data from a particular source (e.g., website), the copyright and terms of service of that source should be provided.
        \item If assets are released, the license, copyright information, and terms of use in the package should be provided. For popular datasets, \url{paperswithcode.com/datasets} has curated licenses for some datasets. Their licensing guide can help determine the license of a dataset.
        \item For existing datasets that are re-packaged, both the original license and the license of the derived asset (if it has changed) should be provided.
        \item If this information is not available online, the authors are encouraged to reach out to the asset's creators.
    \end{itemize}

\item {\bf New Assets}
    \item[] Question: Are new assets introduced in the paper well documented and is the documentation provided alongside the assets?
    \item[] Answer: \answerYes{} 
    \item[] Justification: We will release our code, model and dataset once accepted.
    \item[] Guidelines:
    \begin{itemize}
        \item The answer NA means that the paper does not release new assets.
        \item Researchers should communicate the details of the dataset/code/model as part of their submissions via structured templates. This includes details about training, license, limitations, etc. 
        \item The paper should discuss whether and how consent was obtained from people whose asset is used.
        \item At submission time, remember to anonymize your assets (if applicable). You can either create an anonymized URL or include an anonymized zip file.
    \end{itemize}

\item {\bf Crowdsourcing and Research with Human Subjects}
    \item[] Question: For crowdsourcing experiments and research with human subjects, does the paper include the full text of instructions given to participants and screenshots, if applicable, as well as details about compensation (if any)? 
    \item[] Answer: \answerNA{} 
    \item[] Justification: Our work does not involve crowdsourcing or direct interactions with human subjects.
    \item[] Guidelines:
    \begin{itemize}
        \item The answer NA means that the paper does not involve crowdsourcing nor research with human subjects.
        \item Including this information in the supplemental material is fine, but if the main contribution of the paper involves human subjects, then as much detail as possible should be included in the main paper. 
        \item According to the NeurIPS Code of Ethics, workers involved in data collection, curation, or other labor should be paid at least the minimum wage in the country of the data collector. 
    \end{itemize}

\item {\bf Institutional Review Board (IRB) Approvals or Equivalent for Research with Human Subjects}
    \item[] Question: Does the paper describe potential risks incurred by study participants, whether such risks were disclosed to the subjects, and whether Institutional Review Board (IRB) approvals (or an equivalent approval/review based on the requirements of your country or institution) were obtained?
    \item[] Answer: \answerNA{} 
    \item[] Justification: Our study primarily focuses on image generation and is minimally related to human subjects.
    \item[] Guidelines:
    \begin{itemize}
        \item The answer NA means that the paper does not involve crowdsourcing nor research with human subjects.
        \item Depending on the country in which research is conducted, IRB approval (or equivalent) may be required for any human subjects research. If you obtained IRB approval, you should clearly state this in the paper. 
        \item We recognize that the procedures for this may vary significantly between institutions and locations, and we expect authors to adhere to the NeurIPS Code of Ethics and the guidelines for their institution. 
        \item For initial submissions, do not include any information that would break anonymity (if applicable), such as the institution conducting the review.
    \end{itemize}

\end{enumerate}

\appendix
\section{Dataset Crafting}\label{app:crafting}
\paragraph{Prompts}
The baseball glove waits by the fence.\\
A young athlete breaks in a new baseball glove.\\
A baseball glove rests on the dugout bench.\\
His baseball glove hangs by the door.\\
A baseball glove is left on the fence during practice.\\
A grizzly bear fishes in a rushing river.\\
A bear cub explores the forest with curiosity.\\
A bear catches a fish in the river.\\
A black bear forages for berries in the woods.\\
A bear sniffs the forest floor.\\
A fashionable handbag complements an elegant outfit.\\
A woman carries a stylish handbag on her shoulder.\\
A handbag rests on a cafe table during lunch.\\
A handbag holds essentials for a day of shopping.\\
A handbag adds a pop of color to a monochrome look.\\
A sports ball is caught in a fence.\\
A sports ball lies forgotten under a tree.\\
A sports ball sits on a sandy beach.\\
A sports ball rests on a grassy land.\\
A sports ball stands out on a muddy field.\\
A red stop sign halts traffic at an intersection.\\
A stop sign stands alone on a country road.\\
A stop sign is covered in a layer of snow.\\
A stop sign is adorned with event flyers.\\
A stop sign stands at the entrance to a neighborhood.\\
\paragraph{filtering}
We filter the annotations with a confidence score higher than 0.75 and the label is right.

\section{Detector training}\label{app:training}
We utilized the mmdetection repository~\cite{chen2019mmdetection}. Our approach incorporated Generalized Focal Loss~\cite{li2020generalized} and a Resnet101 architecture~\cite{chen2016deep} expanded to 4$\times$ width to accommodate the 4-channel inputs of the initial noises. We adhered to the COCO dataset configurations provided by the repository. During training, we observed that strong data augmentation techniques significantly impaired performance, leading us to eliminate Flip and Random Resize functions. We conducted our training on four NVIDIA GTX 1080 Ti graphics cards using 17.5k noise samples and report the mean Average Precision (mAP) at 50\% threshold on the 1k validation dataset.
\paragraph{Configs} We utilized the mmdetection repository~\cite{chen2019mmdetection}. Our approach incorporated Generalized Focal Loss~\cite{li2020generalized} and a Resnet101 architecture~\cite{chen2016deep} expanded to 4$\times$ width to accommodate the 4-channel inputs of the initial noises. We adhered to the COCO dataset configurations provided by the repository. During training, we found that strong data augmentation techniques degenerate the performance greatly. During training, we observed that strong data augmentation techniques significantly impaired performance, leading us to eliminate Flip and Random Resize Transormations. We conducted our training on four NVIDIA GTX 1080 Ti graphics cards using 17.5k noise samples and reported the mean Average Precision (mAP) at 50\% threshold on the 1k validation dataset.
\section{Enhanced location diversity by removing trigger patches}:\label{app:diversity}
\paragraph{Prompts we use} 
    The golden sunlight filters through the dense canopy of the forest, casting dappled shadows on the moss-covered ground. \\
    A red bicycle leans against a gnarled oak tree, its wheels slightly caked with mud from the morning’s ride. \\
    Nearby, a picnic table is set with a checkered cloth, and atop it rests a basket filled with fresh fruit and sandwiches. \\
    A frisbee lies forgotten on the grass, a few feet away from a sleeping dog with its fur glistening in the sun.
    In the background, a kite dances in the sky, its bright colors a stark contrast against the blue expanse above. \\
    A laptop is open on the table, displaying vibrant images of nature, momentarily abandoned for the allure of the outdoors. \\
    A baseball glove and ball sit on the bench, remnants of a game played in the spirit of friendly competition. \\
    A traffic cone marks the end of a nearby trail, signaling caution to the cyclists and hikers passing by. \\
    A fire hydrant stands at the edge of the clearing, its red paint chipped but vibrant, a silent guardian of safety. \\
    As the day wanes, the street lights begin to flicker on, their glow adding a soft luminescence to the tranquil scene.
\section{Better prompt following by injecting trigger patches}\label{app:injecting}
\paragraph{Prompts we use}
    a sports ball in the left, \\
    a cow in the left, \\
    an apple in the left, \\
    a bicycle in the left, \\
    a vase in the left \\
    a sports ball in the right, \\
    a cow in the right, \\
    an apple in the right, \\
    a bicycle in the right, \\
    a vase in the right \\
\section{Resources} We use 4 1080 Ti for generating and evaluating.
\end{document}